# Design of EMG-driven Musculoskeletal Model for Volitional Control of a Robotic Ankle Prosthesis

Chinmay Shah[1], Aaron Fleming[2], Varun Nalam[2] and He(Helen) Huang[2], *Senior Member, IEEE*

*Abstract*— Existing robotic lower-limb prostheses use autonomous control to address cyclic, locomotive tasks, but they are inadequate to operate the prosthesis for daily activities that are non-cyclic and unpredictable. To address this challenge, this study aims to design a novel electromyography (EMG)-driven musculoskeletal model for volitional control of a robotic ankle-foot prosthesis. This controller places the user in continuous control of the device, allowing them to freely manipulate the prosthesis behavior at will. The Hill-type muscle model was used to model a dorsiflexor and a plantarflexor, which functioned around a virtual ankle joint. The model parameters were determined by fitting the model prediction to the experimental data collected from an able-bodied subject. EMG signals recorded from ankle agonist and antagonist muscle pair were used to activate the virtual muscle models. This model was validated via offline simulations and real-time prosthesis control. Additionally, the feasibility of the proposed prosthesis control on assisting the user's functional tasks was demonstrated. The present control may further improve the function of robotic prosthesis for supporting versatile activities in individuals with lower-limb amputations.

## I. INTRODUCTION

Lower-limb amputation affects the ability of individuals to perform functional tasks, which severely affects the quality of life[1, 2]. These individuals are prescribed passive prostheses which are incapable of restoring much of the biological function of the joint [3]. To overcome this problem considerable advancements have been made in the field of robotic powered prosthesis enabling the development of lightweight, powered prostheses that are capable of generating similar torque and power as that of an able-bodied joint [3-7]. These devices use a finite state controller where the task being performed is divided into various states and as the user passes through these states the device is commanded to perform the appropriate action. These controllers are autonomous; meaning they do not incorporate any form of human input into the control algorithm. They rely on the interaction of the device with the environment to identify the user's locomotion state and execute a preprogrammed set of instructions to perform the cyclic locomotion task[6]. Thus, new prosthesis behavior (i.e., not already pre-programmed) cannot be generated by the user to adapt to ever-changing environments (e.g., standing at a crowded bus stop, pushing a lawnmower, playing soccer, etc.).

The real world poses many uncertainties present in the environment in the form of terrain variations, obstacles, etc. Such uncertainties require spontaneous responses from the user to adapt to these changes. The lack of active control of the prosthetic ankle joint leads to poorer stability and balance confidence. This lack of balance confidence and absence of human intent from these autonomous controllers has been a hindrance to these prostheses being used in the real world.

EMG control has seen a lot of popularity recently as a method for volitional control of lower limb prostheses [8]. The addition of human intent to these controllers has the potential to allow amputees to modulate their prosthesis mechanics to react to the uncertainties, and volitionally perform non-repetitive tasks. There are two popular methods that use EMG control: Supervisory EMG control and Direct EMG control. Supervisory control is a form of discrete control where a pattern classifier is used to identify human intent by selecting the appropriate type of locomotion mode [9, 10]. This type of control requires a significant amount of training data from the user of the device to train the pattern classifiers. But this method is prone to classification errors, where the selection of the wrong locomotion mode may lead to walking instability in prosthesis users. Continuous EMG control is another method used where the EMG signal is directly related to the force generated by the muscle. This type of continuous volitional control mimics the biological musculoskeletal systems and places the human in direct control making reactionary and non-cyclic tasks possible.

While direct EMG control can restore the ability to adapt to various environments it's unclear whether amputees can coordinate residual muscle activations to reproduce normative ankle joint function. Recent studies have shown that with appropriate training transtibial amputees can regain considerable coordination of voluntary residual muscle activity [11]. These studies have also evaluated the ability of amputee individuals to use this proportional myoelectric control to perform typical daily-life tasks and balancing tasks [11, 12]. They implemented proportional myoelectric control on an ankle prosthesis where the muscle behavior was achieved using pneumatic artificial muscles also known as McKibben actuators. The pneumatic actuators modeled muscle behavior while the proportional EMG relayed the human intent to the system. Though this type of control has shown tremendous success, due to the presence of pneumatic actuators which require compressed air to function, this device has been constrained to laboratory testing only.

The goal of this study was to design an EMG-driven musculoskeletal (MSK) model for continuous volitional control of robotic ankle prosthesis. MSK model-based controllers have been applied to lower limb prosthesis control previously, these virtual muscle models capture the adaptability of the muscle to generate force depending on the state of the muscle, but they lack volitional (or EMG) control from the user. These studies have shown the capability of MSK model-based controllers to adapt to walking speeds, incline decline walking[13], and also stair climbing and stair descent [14]. Though these MSK controllers are more adaptable than other autonomous controllers, they are limited to locomotion tasks only. EMG-driven MSK model has been successfully implemented in upper-limb prosthesis control [15, 16]. However, we are unaware of an EMG-driven MSK controller for motorized lower-limb prosthesis control. Therefore, this study made the following contribution to the field. 1) We developed a new EMG-driven MSK model to determine the ankle joint torque based on EMG signals from the *tibialis anterior* and the *gastrocnemius*. 2) For the first time, we successfully implemented the control on a motorized robotic ankle prosthesis. 3) We showed the feasibility of our proposed control to assist with daily activities that did not include cyclic, predictive motion in the ankle.

*Research supported by NSF 1954587 and NIH NICHD F31HD101285
[1]Chinmay Shah is with the Department of Mechanical and Aerospace Engineering at North Carolina State University, USA (cshah@ncsu.edu).
[2]Aaron Fleming, Varun Nalam and He(Helen) Huang are with the Joint Department of Biomedical Engineering at North Carolina State University and University of North Carolina at Chapel Hill, USA (ajflemin@ncsu.edu, vnalam@ncsu.edu, hhuang11@ncsu.edu)

The rest of the paper was structured in the following way. Section II introduces the methodology of the MSK design and evaluation procedure. Section III describes and discusses the results for the offline simulations and real-time performance evaluation of the EMG-driven MSK model. Section IV discusses the impact and future direction of this study.

## II. METHODOLOGY

### A. Control Architecture

In this paper, we propose an EMG-driven MSK model-based controller where we use the EMG from the *tibialis anterior (TA)* and *medial gastrocnemius (GAS)* to activate a pair of antagonistic virtual muscles, modeled after the antagonistic muscles present around the ankle joint. While several muscles contribute to the ankle flexion moment in the sagittal plane, developing the model that includes all muscles would lead to a highly complex model, requiring many parameters and multiple EMG inputs [15]. Thus, we modeled the virtual muscles as a lumped parameter model, where one muscle was used to represent the combined effect of multiple muscles responsible for generating the required force. We used this virtual muscle model combined with the musculoskeletal geometry to predict the ankle joint torque. During real-time control, this predicted ankle torque was sent as the reference torque command to the *VSeM* robotic ankle prosthesis[17]. Fig. 1 depicts the control layout with the important parts of the EMG-driven MSK model, A. Activation Dynamics B. Musculoskeletal Geometry, and C. Muscle Contraction Dynamics.

### B. Activation Dynamics (EMG to Muscle Activation)

For real-time control, we placed EMG electrodes (Motion Lab Systems, MA 400 ) on the *TA* and *GAS*. We high pass filtered (cutoff at 40Hz, 4th Order Butterworth) to get rid of any motion artifacts. We then rectified the signal before implementing a moving average function with a 100ms sliding window to develop the EMG envelope. We then normalized the resulting signal using the max signal values obtained during the beginning of the session where we asked the participant to flex each muscle maximally three times. We took the max value as the average of the three generated peaks.

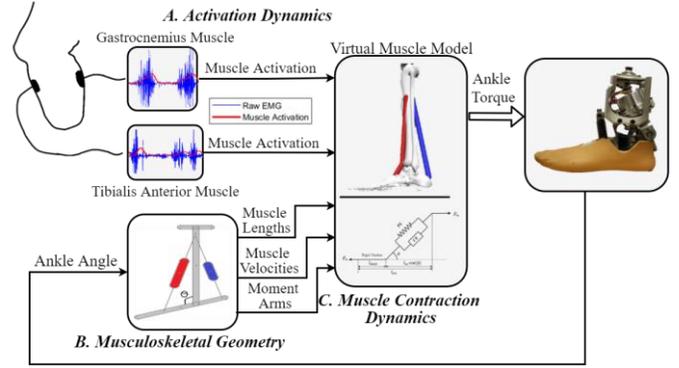

**Figure 1.** EMG-driven MSK model-based controller layout and model components - A. Activation Dynamics (EMG to muscle activation) B. Musculoskeletal Geometry, C. Muscle Contraction Dynamics.

We used this rectified, normalized, and enveloped EMG ($e(t)$) signal to calculate the corresponding neural activation ($u(t)$), using a discrete version of a second-order differential equation used to represent the relation between EMG and neural activation [18],

$$u(t) = \alpha\, e(t-d) - \beta_1\, u(t-1) - \beta_2\, u(t-2) \quad (1)$$

where $\alpha = 0.9486$, $\beta_1 = -0.056$ & $\beta_2 = 0.000627$ [19]. $d$ represents the electromechanical delay present between the onset of the EMG signal and the contraction of the muscle to generate force. This delay was determined based on the experimental data collected for the training of the model and further tuned for real-time control corresponding to the delay present in the robotic ankle prosthesis. At lower levels of force, this neural activation has a non-linear relationship with muscle activation. This relationship was captured by the given equation [19].

$$a(t) = \frac{e^{Au(t)} - 1}{e^A - 1} \quad (2)$$

where A represents the non-linear shape factor, which is constrained between, $-3 < A < 0$. $a(t)$ represents the final input that was sent to the virtual muscle model as muscle activation.

### C. Musculoskeletal Geometry

The musculoskeletal geometry defines how the muscle attaches to the skeleton. It is used to capture the relationship between the change in length of the moment arm with the change in joint angle. The joint torque produced is the product of the muscle force and the moment arm, $\tau_m = r(\theta)F_m$. The moment arm $r$ is a function of joint angle ($\theta$) and this relation was given by the following equation [20].

$$r(\theta) = r_{max}\, cos(\theta - \theta_{max}) \quad (3)$$

where $r_{max}$ is the maximum moment arm length and $\theta_{max}$ is the angle at which, the moment arm is maximum in length. $\theta$ is the ankle joint angle measured between the shank and the foot. This angle was measured using the encoder built into the ankle prosthesis when the control was implemented in real-time on the robotic ankle prosthesis. The length of the muscle unit is also a function of the joint angle, it was computed using the following relation[20].

$$l_{mt} = l_o \cos(\phi_{ref}) + l_{slack} - r\sin(\theta_{max} - \theta_{ref}) + r\sin(\theta_{max} - \theta) \quad (4)$$

where $l_{mt}$ is the length of the muscle unit, which is defined as the length of the entire muscle between the attachment points. $l_o$ is the optimal muscle fiber length, which is the length of the muscle fiber when the muscle generates peak force. $l_{slack}$ represents the slack length of the tendon unit in the muscle model. $\theta_{ref}$, $\phi_{ref}$ are the ankle angle and pennation angle respectively, when the muscle fiber length ($l_{ce}$) is equal to the optimal muscle fiber length ($l_o$).

The length of the muscle model can also be expressed as the sum of contractile element length (muscle fiber length) and series elastic element(tendon) length.

$$l_{mt} = l_{ce} \cos(\phi) + l_{se} \quad (5)$$

Since a rigid tendon model was assumed, the length of the tendon/series elastic element does not change i.e., $l_{slack} = l_{se}$. This affects the accuracy of the model but helps to reduce the complexity of the problem [21].

$$\phi(t) = \sin^{-1}\left(\frac{l_o \sin(\phi_{ref})}{l_{ce}}\right) \quad (6)$$

The pennation angle, ϕ, is not constant and changes with the length of the muscle unit. For Eq. 6 to be valid an assumption that the muscle fibers maintain a constant thickness and volume during muscle contraction was made [19].

Eq 4, 5, and 6 and their derivatives were used to compute $l_{ce}$ and $v_{ce}$. Length of the contractile element ($l_{ce}$) and velocity of contraction ($v_{ce}$) are the inputs sent to the muscle model for calculating the corresponding muscle force.

*D. Muscle Contraction Dynamics*

The Hill-type muscle model consists of an active contractile element (responsible for generating the force), a parallel passive element, and a series elastic element (tendon). As a rigid tendon assumption was made, the series elastic element was not included in the model. The force generated by the muscle model $F_m$ was given by

$$F_m = (F_{ce} + F_{pe})\cos(\phi) \quad (7)$$

where $F_{ce}$ represents the force generated by the contractile element and $F_{pe}$ denotes the force generated by the parallel elastic element. $F_{ce}$ is a function of $l_{ce}$, $v_{ce}$ and $a$ which represent the length of the contractile element, the velocity of contraction, and muscle activation respectively.

$$F_{ce} = F_{max} F_l(l_{ce}) F_v(v_{ce}) a \quad (8)$$

$F_{max}$ represents the maximum isometric force generated by the muscle. $F_l(l_{ce})$ & $F_v(v_{ce})$ represents the force-length and force-velocity relationship for a Hill-type muscle model, which are defined as [22],

$$F_l(l_{ce}) = e^{\frac{-(l_o - l_{ce})^2}{(l_o w)^2}} \quad (9)$$

$$F_v(v_{ce}) = \begin{cases} \frac{v_{max} - v_{ce}}{v_{max} + K v_{ce}}, v_{ce} < 0 \\ N + (N-1)\frac{v_{max} + v_{ce}}{7.56K v_{ce} - v_{max}}, v_{ce} \geq 0 \end{cases} \quad (10)$$

w was fixed to a value of 0.56 [22], this parameter adjusts the width of the bell-shaped curve representing the force-length behavior of the muscle model. In Eq 9. $v_{max}$ represents the maximum contraction velocity of the muscle fiber which was set to $10 \frac{l_o}{s}$ [23, 24]. K is the shape parameter fixed to 5 [22] and N is a dimensionless force constant fixed at 1.5 [22]. All these parameters define the dynamic force-length and force-velocity relations of the muscle. The parallel elastic element present in the system only contributes to force generation if the muscle fibers are stretched beyond the optimal fiber length $l_o$ [22].

$$F_{pe} = F_{max}\left(\frac{l_{ce} - l_o}{l_o \epsilon_{pe}}\right)^2, l_{ce} > l_o \quad (11)$$

In Eq 11. $\epsilon_{pe}$ is the reference strain which is fixed at 0.56 [22]. Combining all of the equations, the net ankle joint torque was computed by summing the torque contribution of the 2 muscles present in the model,

$$\tau_{ankle} = F_{m(dorsi)} r_{dorsi}(\theta) - F_{m(plant)} r_{plant}(\theta) \quad (12)$$

where $\tau_{ankle}$ represents the model predicted ankle torque.

*E. Model Parameter Optimization*

The EMG-driven MSK model consists of many parameters that allow it to replicate the dynamic force-generating behavior of muscles. While some parameters are fixed to preserve this dynamic relationship, some parameters are responsible for scaling the muscle force according to a specific muscle in the body. The optimization problem was set up to find a set of parameters that would allow the muscle model predictions to best fit the biological ankle torque collected during walking tasks. $F_{max}$, $l_o$, $r_{max}$, $\theta_{ref}$, $\theta_{max}$ were the physiological parameters chosen for optimization. $\theta_{ref}$ and $\theta_{max}$ values were shared by both the muscles; where their function was interchanged i.e., $\theta_{ref}$ for dorsiflexor muscle was equal to $\theta_{max}$ of the plantar flexor muscle and vice versa. This was done to account for the orientation of the muscles and to reduce the total number of parameters chosen for optimization.

MATLAB's built-in *GlobalSearch* function was recruited to perform the optimization[15]. The goal of this optimization function was to minimize the sum of squared error between the model predicted ankle torque $\tau_{ankle}$ and biological ankle torque $\tau_{bio}$, where the error was given by,

$$error = \frac{1}{n}\sum_1^n (\tau_{ankle} - \tau_{bio})^2 \qquad (13)$$

n is the length of data used in optimization. Table I. depicts the allowable range of the parameters for optimization and the parameter values obtained after optimization.

TABLE I.  NUMERICAL OPTIMIZATION CONSTRAINTS AND OPTIMIZED PARAMETER VALUE

| Model Parameter | Min Value | Max Value | Fitted Value |
|---|---|---|---|
| **Plantar flexor** | | | |
| $F_{max}$ (N) | 500 | 6000 | 4800 |
| $l_o$ (m) | 0.02 | 0.06 | 0.0402 |
| $r_{max}$ (m) | 0.01 | 0.065 | 0.0375 |
| $\theta_{ref}$ (deg) | 70 | 130 | 70 |
| $\theta_{max}$ (deg) | 70 | 130 | 112 |
| **Dorsi flexor** | | | |
| $F_{max}$ (N) | 500 | 4000 | 1800 |
| $l_o$ (m) | 0.02 | 0.145 | 0.065 |
| $r_{max}$ (m) | 0.01 | 0.065 | 0.0449 |

*F. Evaluation Protocol*

We recruited one able-bodied participant (175cm height and 75kg weight) for the study with IRB approval and signed informed consent. To build and validate a subject-specific model, bipolar EMG electrodes were placed on the *TA* and *GAS*. The muscles were located by palpation and visualizing the EMG signal before the data collection. To capture the kinematics of the subject, retroreflective markers were placed on the subject. A 12-camera motion capture system (Vicon, UK) was used to track the marker positions at 100 Hz. The ground reaction forces were measured at 1000 Hz using a split-belt instrumented treadmill (Bertec, USA). The participant was asked to perform various tasks such as walking on the treadmill at 1.0m/s, 1.2m/s, and 1.4m/s, squatting, standing on their toes, and performing sit-to-stand transitions during data collection. Inverse dynamics was performed using the collected data to calculate the torque generated at the ankle joint when these tasks were performed. EMG signals, ankle angle, and calculated ankle torque during walking tasks were used to establish the MSK model parameters via optimization as discussed in the previous section.

Evaluation of our proposed EMG-driven MSK model control included three parts. In the first part, we conducted an offline evaluation of the EMG-driven MSK model for predicting the ankle joint torque. Trials that were not involved in the optimization procedure for building the MSK model were used to evaluate the model. Next, the EMG-driven model-based controller was validated in real-time. The human ankle was attached to an able-bodied adaptor to allow able-bodied subjects to wear and operate the device(as shown in the attached supplementary video). Due to the inability of the subject to activate the *GAS* and *TA* individually when the foot was fixed in the adaptor, the EMG signals were recorded from the intact side. This restricted testing of the controller to only bilaterally symmetrical tasks. In the second part of the evaluation, we tested the model-based control in an open loop with non-weight bearing posture tasks. The subject sat on a chair comfortably and performed ankle dorsiflexion and plantarflexion repeatedly on the intact limb. The EMG-driven MSK model estimated the intact ankle torque and then applied it to the prosthetic ankle. In the third part of the evaluation, we aimed to show the feasibility of our proposed control in a closed-loop (with a human operator in the loop), using weight-bearing tasks. The tasks included standing on toes and performing sit-to-stand transitions. The measurement methods during these real-time evaluations were the same as the measurements when collecting data for training and offline validation of the model. The prosthesis-generated ankle torque was measured using a loadcell present in the ankle prosthesis[17].

*H. Data Analysis and Evaluation Metrics*

The optimization was performed in MATLAB (Mathworks, Natick, MA). Biological ankle torque was calculated using the kinematic joint motion and ground reaction forces using an inverse dynamic model. Visual 3D (C-motion Inc., USA) was used to perform the inverse dynamics calculations.

In the first part of the evaluation, we used $R^2$ and Normalized Root Mean Square Error (NRMSE) values between inverse-dynamic calculated biological ankle torque and model predicted ankle torque to assess the accuracy of the MSK model. Since all the tested tasks were repeated motions, we segmented each repetition, normalized the duration of each repetition as % completion of each attempt, and averaged all the repetitions across the same task. In the second and third parts of the evaluation, NRMSE and $R^2$ values were calculated between the model predicted ankle torque, the ankle angle/torque of the prosthesis, and biological ankle angle/torque on the intact side.

### III. RESULTS AND DISCUSSION

*A. Offline Simulations*

Fig 2. compares the ankle torque predicted by the EMG-driven MSK (blue dashed line) and the biological ankle torque calculated by the inverse dynamics (red solid line) for various tasks. Table II. lists the NRMSE and $R^2$ values between measured and predicted ankle torque.

For walking at different speeds (1.0m/s and 1.4m/s) and sit-to-stand transitions we observed similar results between our offline analysis and biological ankle torque ( Fig 2 A., B. C.). The prediction accuracy in the case of standing on toes and squatting was relatively low which can be seen from Fig 2 D. & E. and the $R^2$ and NRMSE values in Table II. This reduced prediction accuracy in some tasks might be because the model was trained on the experimental data collected during the walking task and it couldn't adapt very well to different tasks that have a different range of motion and torque requirements.

**TABLE II.** ANKLE TORQUE PREDICTION ACCURACY DURING OFFLINE SIMULATIONS VIA EMG-DRIVEN MSK MODEL.

| *Task* | *NRMSE* | *$R^2$* |
|---|---|---|
| *Walking 1m/s* | 0.1173 | 0.8210 |
| *Walking 1.4 m/s* | 0.0935 | 0.8895 |
| *Sit to stand* | 0.1695 | 0.7792 |
| *Standing on Toes* | 0.1656 | 0.6842 |
| *Squatting* | 0.1882 | 0.6205 |

*B. Real-Time Open-Loop Control in Non-Weight Bearing Posture Task*

When the MSK model is used to control prosthesis in an open loop, in general, the robotic prosthesis ankle angle can follow the intact joint angle (Fig. 3C.). Both motions showed qualitative agreement without the presence of significant delay between the prosthesis side movement and intact side. ($R^2$ 0.7749 NRMSE 0.1935). The range of motion of the device, however, was lower than the intact side (Fig 3C.). This was due to the limitations of the device used in testing.

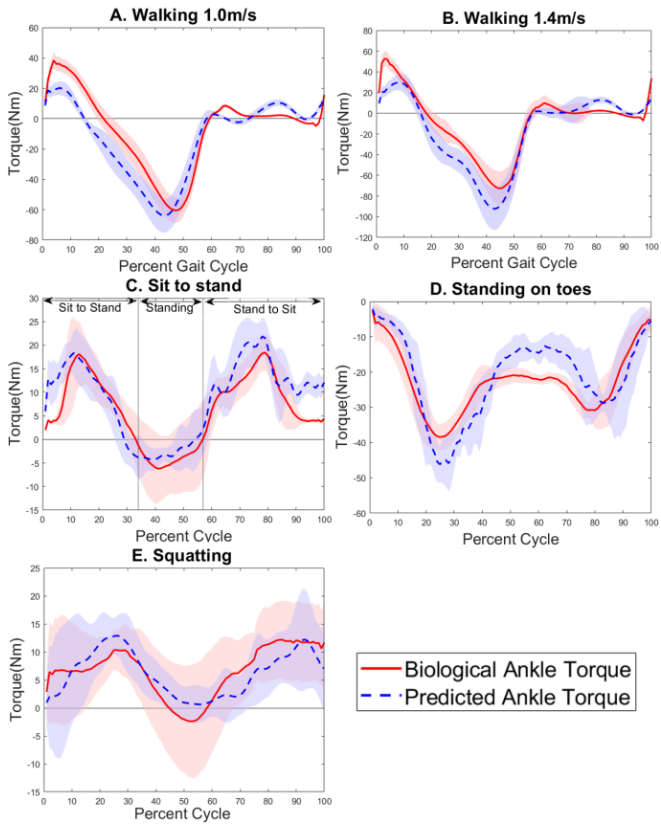

**Figure 2.** Comparison of the biological ankle torque calculated by inverse dynamics (+ve torque corresponds to dorsiflexion torque and -ve torque represents the plantarflexion torque) with MSK model predicted torque for different tasks: A. walking at 1.0 m/s, B. walking at 1.4 m/s, C. sit-to-stand transitions, D. standing on toes, E. squatting. The solid and dash data represent the averaged curve across multiple repetitions. Shaded area denotes +/- one standard deviation.

This control is capable of allowing the user to freely manipulate the ankle position. This is a particularly useful ability that can allow the users to adjust the ankle angle according to the environment, potentially allowing for better obstacle avoidance, sitting posture, terrain adaptation, etc.

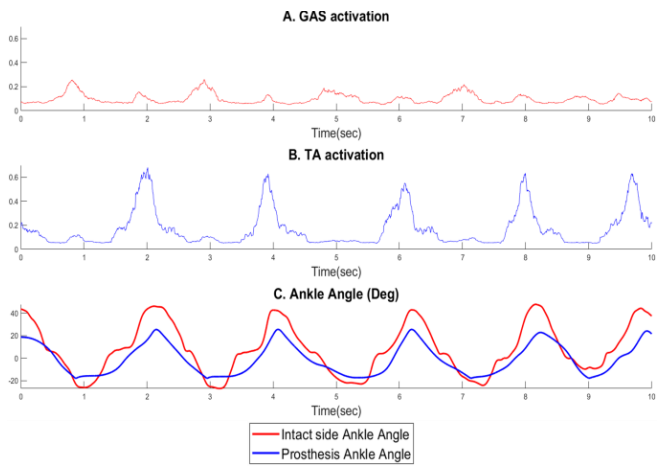

**Figure 3.** Real-time performance of EMG-driven MSK model-based control in open loop. *A*: Muscle activation of *GAS* on the intact side; *B*: Muscle activation of *TA* on the intact side. They were the inputs of the control. *C*: Prosthesis ankle angle was the output of the control. It was compared to the intact ankle angle (+ve values-dorsiflexion and -ve values-plantarflexion).

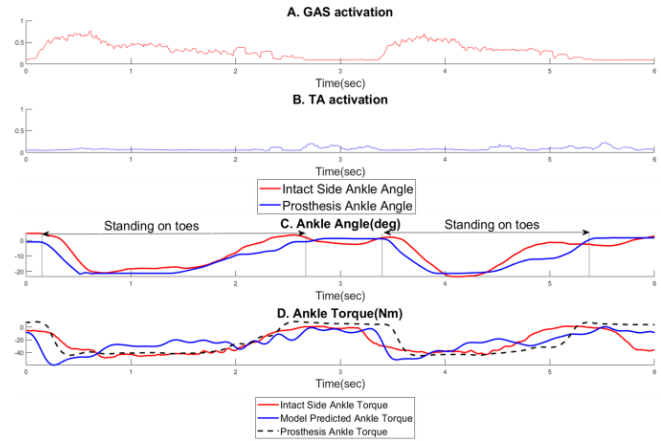

**Figure 4.** Real-time performance for Standing on toes task. A. and B. Muscle activation for *GAS* and *TA* muscles, respectively. C. Comparison of the ankle angle for the intact and prosthetic side (+ve values – dorsiflexion, -ve values – plantarflexion). D. Comparison of ankle torque of the intact side, the model predicted torque and the torque generated by the prosthetic ankle. (+ve values – dorsiflexion torque, -ve values – plantarflexion torque).

### C. Real-Time Closed-loop Control in Weight-Bearing Symmetric Tasks

The two tasks that were chosen were - Standing on toes and sit-to-stand transitions. Fig -4 and Fig-5 depict the real-time performance of the control when tasks were performed. Table III. compares the NRMSE and $R^2$ values for averaged torque and angles values for the different conditions tested.

Sit-to-stand transitions and standing on toes were tested as they are volitional, non-repetitive tasks, that other autonomous controllers cannot perform or assist with. The model predicted torque showed a high correlation with the intact side torque and the torque generated by the prosthesis, which can be seen in Table III (Real-time control with the device is shown in the attached supplementary video). Interestingly, in both cases, the accuracy of the MSK model in ankle torque prediction here was higher than the accuracy shown in offline analysis (Table II). Though EMG signals used to generate prosthetic ankle torque were obtained from the intact limb, the improvement in accuracy may be a result of differences between intact and prosthetic limb states (i.e., the joint angles were not identical). In the future, we plan to shift testing with amputee subjects where their residual muscles will be used to drive the robotic prosthetic ankle torque. In that case, even though the model estimation has errors, the human operator can modify the muscle contraction to meet the task goal.

### IV. CONCLUSION

In this study, we developed an EMG-driven MSK model-based controller for real-time control of a motorized, robotic ankle prosthesis. The model prediction accuracy was evaluated through offline simulations, where the model predicted torque was compared with the biological ankle torque, computed via inverse dynamics. The EMG-driven MSK model was also implemented on a robotic ankle prosthesis and evaluated online. The results showed that the personalized MSK model can accurately predict the ankle joint torque in various task contexts based on the activity of antagonistic ankle muscles. In addition, for the first time, we showed the feasibility of design of robotic ankle control based on biomimicry principles to enable the performance of prosthesis users on tasks that are unpredictable and non-cyclic.

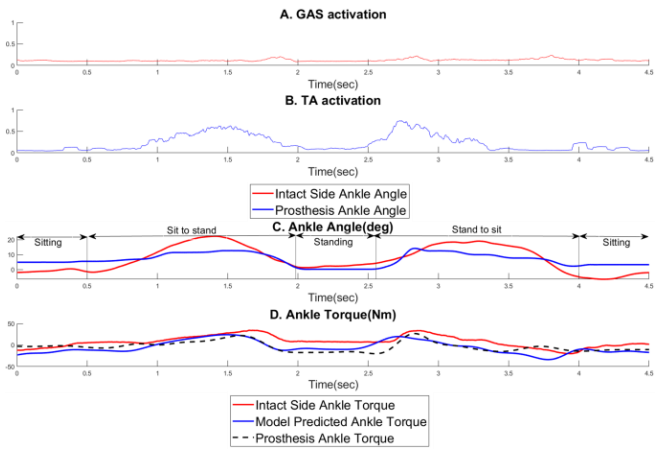

**Figure 5.** Real-time performance for sit-to-stand transitions. A. and B. Muscle activation for *GAS* and *TA* respectively. C. Comparison of the ankle angle for the intact and prosthetic side (+ve values – dorsiflexion, -ve values – plantarflexion). D. Comparison of ankle torque of the intact side, the model predicted torque and the torque generated by the prosthetic ankle. (+ve values – dorsiflexion Torque, -ve values – plantarflexion Torque).

**TABLE III.** REAL-TIME PERFORMANCE OF EMG-DRIVEN MSK MODEL DURING WEIGHT-BEARING TASKS

| Standing on toes task | NRMSE | $R^2$ |
|---|---|---|
| Ankle angle vs Prosthesis angle | 0.1461 | 0.9572 |
| Model predicted torque vs Intact side torque | 0.2223 | 0.8671 |
| Model predicted torque vs Prosthesis torque | 0.1044 | 0.9689 |
| **Sit to stand task** | **NRMSE** | **$R^2$** |
| Ankle angle vs Prosthesis angle | 0.1752 | 0.6071 |
| Model predicted torque vs Intact side torque | 0.41 | 0.7012 |
| Model predicted torque vs Prosthesis torque | 0.1287 | 0.8921 |

This study also had limitations. As the evaluation was performed with an able-bodied subject using the able-bodied adaptor, the real-time tests were restricted to only bilaterally symmetric tasks. Our future efforts will focus on testing the proposed control on individuals with transtibial amputations when performing more activities in daily living.

ACKNOWLEDGMENT

The authors would like to thank Dr. Ming Liu, Wentao Liu, and Sameer Upadhye for their support in this work.